\documentclass{article}

\usepackage{arxiv}

\usepackage[utf8]{inputenc} 
\usepackage[T1]{fontenc}    
\usepackage{hyperref}       
\usepackage{url}            
\usepackage{booktabs}       
\usepackage{amsfonts}       
\usepackage{amsmath}
\usepackage{nicefrac}       
\usepackage{microtype}      
\usepackage{lipsum}

\usepackage{graphicx}
\usepackage{tabularx}
\usepackage{multirow}
\newenvironment{methods}{}

\title{Evaluating recommender systems for AI-driven biomedical informatics}

\author{
    William La Cava, Heather Williams, Weixuan Fu, Steve Vitale, Durga Srivatsan, Jason H. Moore\thanks{Corresponding author: \texttt{jhmoore@upenn.edu}}\\
  Institute for Biomedical Informatics\\
  Department of Biostatistics, Epidemiology and Informatics\\
  University of Pennsylvania\\
  Philadelphia, PA 19104 \\
}

\begin{document}

\maketitle

\textbf{Motivation:}
Many researchers with domain expertise are unable to easily apply machine learning to their bioinformatics data due to a lack of machine learning and/or coding expertise. 
Methods that have been proposed thus far to automate machine learning mostly require programming experience as well as expert knowledge to tune and apply the algorithms correctly. 
Here, we study a method of automating biomedical data science using a web-based platform that uses AI to recommend model choices and conduct experiments. 
We have two goals in mind: first, to make it easy to construct sophisticated models of biomedical processes; and second, to provide a fully automated AI agent that can choose and conduct promising experiments for the user, based on the user's experiments as well as prior knowledge. 
To validate this framework, we experiment with hundreds of classification problems, comparing to state-of-the-art, automated approaches. 
Finally, we use this tool to develop predictive models of septic shock in critical care patients. 
\\
\textbf{Results:} 
We find that matrix factorization-based recommendation systems outperform meta-learning methods for automating machine learning. 
This result mirrors the results of earlier recommender systems research in other domains. 
The proposed AI is competitive with state-of-the-art automated machine learning methods in terms of choosing optimal algorithm configurations for datasets. 
In our application to prediction of septic shock, the AI-driven analysis produces a competent machine learning model (AUROC 0.85 +/- 0.02) that performs on par with state-of-the-art deep learning results for this task, with much less computational effort.

\textbf{Availability:} PennAI is available free of charge and open-source. It is distributed under the GNU public license (GPL) version 3.\\
\textbf{Contact:} \href{lacava@upenn.edu}{lacava@upenn.edu}\\
\textbf{Supplementary information:} Software and experiments are available from \href{https://epistasislab.github.io/pennai}{epistasislab.github.io/pennai}.


\def\mycolwidth{0.5\textwidth}
\section{Introduction}
\label{s:introduction}

Experimental data is being collected faster than it can be understood across scientific disciplines~\cite{bansal_big_2014}. 
The hope of many in the data science community is that widely accessible, open-source artificial intelligence (AI) tools will allow scientific insights from these data to keep abreast of their collection~\cite{olson_system_2017}. 
AI is expected to make significant improvements to scientific discovery, human health and other fields in the coming years. 
One of the key promises of AI is the automation of learning from large sets of collected data. 
However, at the same time that data collection is outpacing researchers, methodological improvements from the machine learning (ML) and AI communities are outpacing their dissemination to other fields.
As a result, AI and ML remain steep learning curves for non-experts, especially for researchers pressed to gain expertise in their own disciplines. 

Specialized researchers would benefit greatly from increasingly automated, accessible and open-source tools for AI. 
With this in mind, we created a free and open-source platform called PennAI\footnote{\url{http://github.com/EpistasisLab/pennai}} (University of Pennsylvania Artificial Intelligence) that allows the non-expert to quickly conduct a ML analysis on their data~\cite{olson_system_2017}. 
PennAI uses a web browser-based user interface (UI) to display a user's datasets, experiments, and results, shown in Fig.~\ref{fig:ui}.
In order to automate the user's analysis, PennAI uses a bootstrapped recommendation system that automatically configures and runs supervised learning algorithms catered to the user's datasets and previous results. 

In addition to its use as a data science tool, PennAI serves as a test-bed for methods development by AI researchers who wish to develop automated machine learning (AutoML) methods.  
AutoML is a burgeoning area of research in the ML community that seeks to automatically configure and run learning algorithms with minimal human intervention. 
A number of different learning paradigms have been applied to this task, and tools are available to the research community~\cite{hutter_sequential_2011,kotthoff_auto-weka_2017,feurer_efficient_2015,komer_hyperopt-sklearn:_2014,olson_evaluation_2016,feurer_practical_2018} as well as commercially\footnote{\href{http://h20.ai}{H2O}}~\cite{real_using_2018}.
A competition around this goal has been running since 2015\footnote{\url{http://automl.chalearn.org/}} focused various budget-limited tasks for supervised learning~\cite{guyon_brief_2016}.

Despite what their name implies, many leading AutoML tools are highly configurable and require coding expertise to operate~\cite{hutter_sequential_2011,kotthoff_auto-weka_2017,feurer_efficient_2015,komer_hyperopt-sklearn:_2014,olson_evaluation_2016,feurer_practical_2018}, which leaves a gap for adoption to new users.
Furthermore, AutoML tools typically wrap many ML analyses via the automation procedure, obscuring the analysis from the user and preventing user intput or guidance. 
In contrast to these strategies, PennAI interacts with the user through algorithm recommendations, and can learn from both its own analysis and that of the user. 
This interaction is achieved using a web-based recommender system.

Recommender systems are well known inference methods underlying many commercial content platforms, including Netflix~\cite{bennett_netflix_2007}, Amazon~\cite{smith_two_2017}, Youtube~\cite{davidson_youtube_2010}, and others.  
There has been limited adoption of recommender systems as AutoML approaches~\cite{yang_oboe:_2018,fusi_probabilistic_2018}, much less an attempt to compare and contrast different underlying strategies. 
To fill this gap, we benchmark several state-of-the-art recommender systems in a large experiment that tests their use as an AutoML strategy. 

Our first goal is to assess the ability of state-of-the-art recommender systems to learn the best ML algorithm and parameter settings for a given dataset over a set number of iterations in the context of previous results. 
We compare collaborative filtering (CF) approaches as well as metalearning approaches on a set of 165 open-source classification problems. 
We find that the best algorithms (matrix factorization) work well without leveraging dataset metadata, in contrast to other AutoML approaches.
We demonstrate the ability of PennAI to outperform Hyperopt and perform on par with AutoSklearn, two popular AutoML tools for Python.
Our second goal is to test PennAI in application to predictive modeling in the biomedical context. 
To this end, we use PennAI to develop predictive models of septic shock using the MIMIC-III~\cite{johnson_mimic-iii_2016} critical care database.
We find that PennAI is useful for quickly finding models with strong performance, in this instance producing a model with performance on par with complex, effort-intensive recurrent deep neural networks.
We include a sensitivity analysis of the predictive model of septic shock produced by PennAI that lends credence to its predictions. 

\section{Background}
\label{s:background}

In this section, we briefly review AutoML methodologies, which are a key component to making data science approachable to new users.
We then describe recommender systems, the various methods that have worked in other application areas, and our motivation for applying them to this relatively new area of AutoML. 

\subsection{Automated Machine Learning}

AutoML is a burgeoning area of research in the ML community that seeks to automatically configure and run learning algorithms without human intervention. 
A number of different learning paradigms have been applied to this task, and tools are available to the research community as well as commercially. 
A competition around this goal has been running since 2015~\footnote{\url{http://automl.chalearn.org/}} focused various budget-limited tasks for supervised learning~\cite{guyon_brief_2016}.

A popular approach arising from the early competitions is sequential model-based optimization via Bayesian learning~\cite{hutter_sequential_2011}, represented by the auto-Weka, AutoSklearn and Hyperopt packages \cite{kotthoff_auto-weka_2017,feurer_efficient_2015,komer_hyperopt-sklearn:_2014}.
These tools parameterize the combined problem of algorithm selection and hyperparameter tuning and use Bayesian optimization to select and optimize algorithm configurations.

Auto-sklearn incorporates metalearning into the optimization process \cite{brazdil_ranking_2003,brazdil_metalearning:_2008} to narrow the search space of the optimization process.
Metalearning in this context refers to the use of the ``metafeatures" of the datasets, such as predictor distributions, variable types, cardinality etc. provide information about algorithm performance that can be leveraged to choose an appropriate algorithm configuration, given these properties for a candidate dataset.
Auto-sklearn uses metalearning to narrow the search space of their learning algorithm. 
In lieu of metalearning, PoSH AutoSklearn~\cite{feurer_practical_2018}, an update to AutoSklearn, opted to bootstrap AutoSklearn with an extensive analysis to minimize the configuration space. 
ML configurations were optimized on a large number of datasets beforehand, and the initial configurations were narrowed to those that performed best over all datasets. 
This tool effectively replaced metalearning with bootstrapping; our experiments in Section~\ref{s:experiments} provide some evidence supporting a similar strategy.

Another popular method for AutoML is tree-based pipeline optimization tool TPOT~\cite{olson_evaluation_2016}. 
TPOT uses an evolutionary computation approach known as genetic programming to optimize syntax tree representations of ML pipelines. 
Complexity is controlled via multi-objective search. 
Benchmark comparisons of TPOT and AutoSklearn show trade-offs in performance for each~\cite{balaji_benchmarking_2018}. 

There are many commercial tools providing variants of AutoML as well. 
Many of these platforms do not focus on choosing from several ML algorithms, but instead provide automated ways of tuning specific ones. 
Google has created AutoML tools as well using neural architecture search~\cite{real_using_2018}, a method for configuring the architecture of neural networks.
This reflects their historical focus on learning from sequential and structured data like images. 
\href{http://h20.ai}{H2O} uses genetic algorithms to tune the feature engineering pipeline of a user-chosen model. 
Intel has focused on proprietary gradient boosted ensembles of decision trees~\cite{guyon_brief_2016}. 

A main paradigm of many AutoML is that they wrap several ML analyses and return a single (perhaps ensemble) result, thereby obscuring their analysis from the user. 
Although this does indeed automate the ML process, it removes the user from the experience. 
In contrast to these strategies, PennAI uses a recommender system as its basis of algorithm recommendation with the goal of providing a more intuitive and actionable user experience. 

There has a line of research utilizing recommender systems for algorithm selection~\cite{stern_collaborative_2010,misir_alors:_2017,fusi_probabilistic_2018,yang_oboe:_2018}. 
One recent example is Yang et al.~\cite{yang_oboe:_2018}, who found that non-negative matrix factorization could be competitive with AutoSklearn for classification. 
A recent workshop\footnote{\url{http://amir-workshop.org/}} also solicited discussion of algorithm selection and recommender systems, although most research of this nature is interested in tuning the recommendation algorithms themselves~\cite{cunha_metalearning_2018}. 

Ultimately, the best algorithm for a dataset is highly subjective: a user must balance their wants and needs, including the accuracy of the model, its interpretability, the training budget and so forth.
PennAI's coupling of the recommendation system approach with the UI allows for more user interaction, essentially by maintaining their ability to ``look under the hood".
The user is able to fully interface with any and all experiments initialized by the AI in order to, for example, interrupt them, generate new recommendations, download reproducible code or extract fitted models and their results. 
In addition, by making it easy to launch user-chosen experiments, PennAI makes it possible to train the recommender on user-generated results, in addition to its own.
A summary of the differences in features between PennAI and several other AutoML tools is shown in Table~\ref{tbl:compare}.
By developing PennAI as a free and open-source tool, we also hope to contribute an extensible research platform for the ML and AutoML communities. 
The code is developed on Github and documents a base recommender class that can be written to the specification of any learning algorithm.
We therefore hope that it will serve as a framework for bring real world users into contact with cutting edge methodologies. 

\subsection{Recommender Systems}
Recommender systems are typically used to recommend \textit{items}, e.g. movies, books, or other products, to \textit{users}, i.e. customers, based on a collection of user ratings of items. 
The most popular approach to recommender systems is collaborative filtering (CF). 
CF approaches rely on user ratings of items to learn the explicit relationship between similar users and items. 
In general, CF approaches attempt to group similar users and/or group similar items, and then to recommend similar items to similar users. 
CF approaches assume, for the most part, that these similarity groupings are implicit in the ratings that users give to items, and therefore can be learned. 
However, they may be extended to incorporate additional attributes of users or items~\cite{ungar_clustering_1998}.

Recommenders face challenges when deployed to new users, or in our case, datasets. 
The new user \textit{cold start problem}~\cite{schein_methods_2002} refers to the difficulty in predicting ratings for a user with no data by which to group them. 
With datasets, one approach to this problem is through metalearning. 
Each dataset has quantifiable traits that can be leveraged to perform similarity comparisons without relying on algorithm performance history. 
In our experiments we benchmark a recommender that uses metafeatures to derive similarity scores for recommendations, as has been proposed in previous AutoML work~\cite{feurer_efficient_2015, brazdil_metalearning:_2008}. 

Recommender systems are typically used for different applications than AutoML, and therefore the motivations behind different methods and evaluation strategies are also different. 
For example, unlike typical product-based recommendation systems, the AI automatically runs the chosen algorithm configurations, and therefore receives more immediate feedback on its performance. 
Since the feedback is explicitly the performance of the ML choice on the given dataset, the ratings/scores are reproducible, less noisy, and less sparse than user-driven systems. 
This robustness allows us to measure the performance of each recommendation strategy reliably in varying training contexts. 
As another example, many researchers have found in product recommendation that the presence or absence of a rating may hold more weight than the rating itself, since users choose to rate or to not rate certain products for non-random reasons~\cite{marlin_collaborative_2009}. 
This observation has led to the rise of implicit rating-based systems, such as SVD++~\cite{koren_factorization_2008}, that put more weight on presence/absence of ratings. 
In the context of AutoML, it is less likely that the presence of results for a given algorithm configuration imply that it will outperform others.  
Furthermore, the goal of advertising-based, commercial recommendation systems may not be to find the best rating for a user and product, but to promote engagement of the user, vis-a-vis their time spent browsing the website. 
To this end, recommender systems such as Spotlight~\cite{kula_spotlight_2017} are based on the notion of sequence modeling: what is the likelihood of a user interacting with each new content given the sequence of items they have viewed? 
Sequence-based recommendations may improve the user experience with a data science tool, but we contend that they do not align well with the goals of an approachable data science assistant. 

\begin{table*} 
    \centering
    \caption{A comparison of AutoML tool characteristics and PennAI.}
    \label{tbl:compare}
    \begin{tabular}{l r r r r r}
        Tool                &   Methodology             &   Free                &   Open Source     &   Code-Free       &   Learns from User     \\ \midrule
        PennAI              &   Recommender Systems     &   \checkmark          &   \checkmark      &   \checkmark      &   \checkmark           \\
        HyperOptSklearn     &   Bayesian                &   \checkmark          &   \checkmark      &                   &                      \\
        AutoSklearn         &   Bayesian + Metalearning &   \checkmark          &   \checkmark      &                   &                       \\
        TPOT                &   Genetic Programming     &   \checkmark          &   \checkmark      &                   &                      \\
        H20.AI              &   Genetic Algorithms      &                       &   \checkmark      &   \checkmark      &                      \\ \bottomrule
    \end{tabular}
\end{table*}

\begin{methods}
\section{Methods}
\label{s:methods}


\begin{figure*}[ht!]
    \centering
    \includegraphics[height=0.5\textheight]{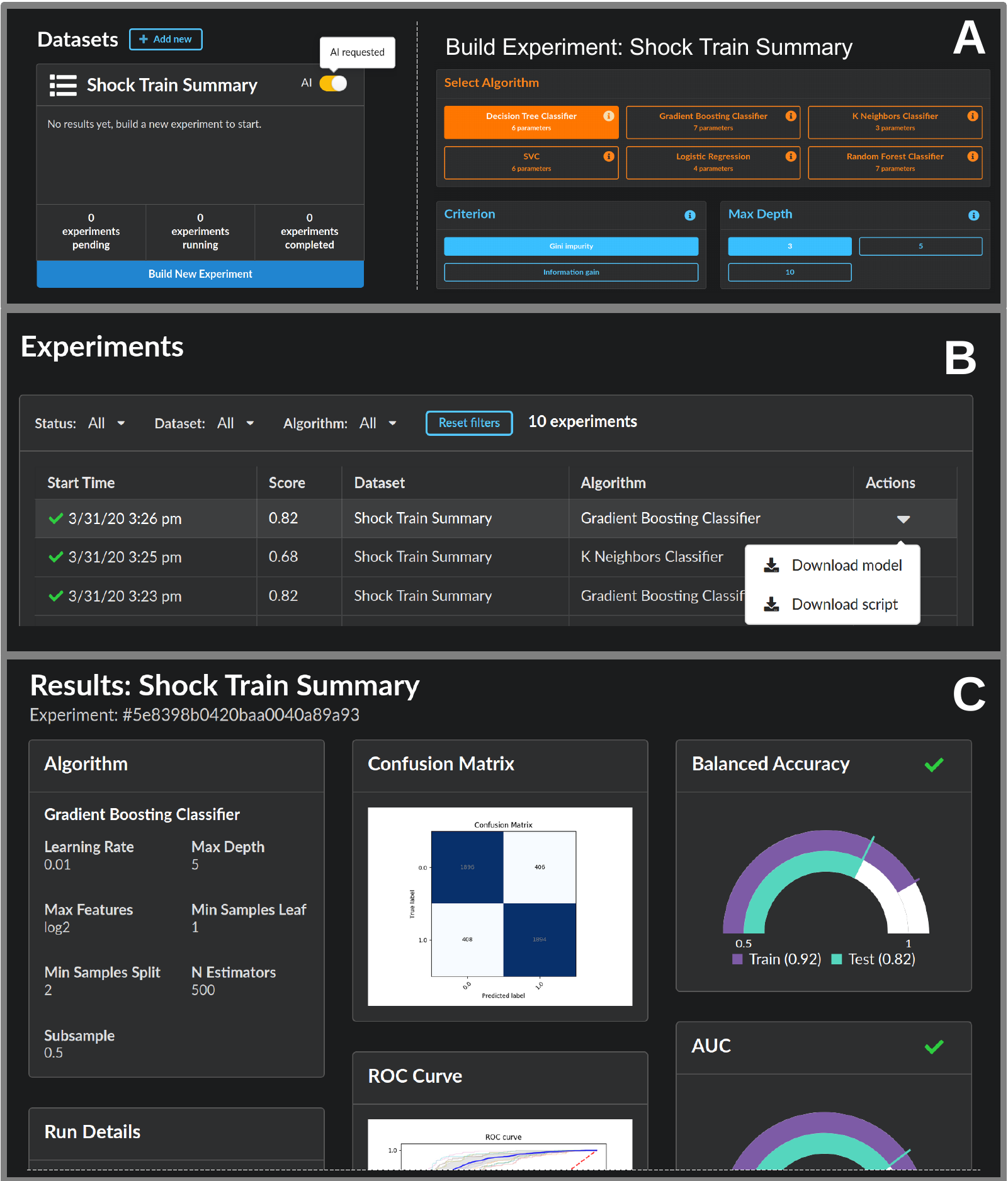}
    \caption{Overview of the UI. 
        A) Users upload datasets and choose a custom experiment (right), or allow the AI to run experiments of its choosing by clicking the AI button (left). 
    B) Experiments are tabulated with configuration and performance information. 
    The user may download scripts to reproduce the experiment in python, or export the fitted model. 
C) The results page displays experiment information and statistics of the fitted model, including various performance measures (confusion matrix, receiver operating characteristic (ROC) curve, etc.) as well as feature importance scores for the independent variables. }
    \label{fig:ui}
\end{figure*}

Here we will briefly describe the user interface of PennAI. 
In Section~\ref{s:methods:rs}, we describe the recommender systems that we benchmark in our experiments for automating the algorithm selection problem.
Fig.~\ref{fig:ui} gives an overview of the data science pipeline. 
Users upload datasets through the interface or optionally by pointing to a path during startup.
At that point, users can choose between building a custom experiment (manually configuring an algorithm of their choice) or simply clicking the AI button. 
Once the AI is requested, the recommendation engine chooses a set of analyses to run.
The AI can be configured with different termination criteria, including a fixed number of runs, a time limit, or running until the user turns it off. 
As soon as the runs have finished, the user may navigate to the results page, where several visualizations of model performance are available (Fig.~\ref{fig:ui}.C)). 

PennAI is available as a docker image that may be run locally or on distributed hardware. 
Due to its container-based architecture, it is straightforward to run analysis in parallel, both for datasets and algorithms, by configuring the docker environment. 
For more information on the system architecture, refer to the Supplemental Material.

\subsection{Recommendation System}
\label{s:methods:rs}

In order to use recommender systems as a data science assistant, we treat datasets as users, and algorithms as items. 
The goal of the AI is therefore as follows: given a dataset, recommend an algorithm configuration to run. 
Once the algorithm configuration has been run, the result is now available as a rating of that algorithm configuration on that dataset, as if the user had rated the item. 
This is a nice situation for recommender systems, since normally users only occasionally rate the items they are recommended.
We denote this knowledge base of experimental results as $\mathcal{D} = \{r_{ad},r_{be},\dots\}$, where $r_{ad}$ is the test score, i.e. rating, of algorithm configuration $a$ on dataset $d$. 
In our experiments the test score is the average 10-fold CV score of the algorithm on the dataset.

With a few notable exceptions discussed below, the recommenders follow this basic procedure:
\begin{enumerate}
    \item Whenever new experiment results are added to $\mathcal{D}$, update an internal model mapping datasets $d$ to algorithm configurations, $a$. 
    \item Given a new recommendation request, generate $\hat{r}_{ad}$, the estimated the score of algorithm configuration $a$ on dataset $d$. 
        Do this for every $ad$ pair that has not already been recommended.
    \item Return recommended algorithm configurations in accordance with the termination criterion, in order of best rating to worst. 
\end{enumerate}
 
Note that the knowledge base $\mathcal{D}$ can be populated not only by the AI, but by the user through manual experiments (Fig.~\ref{fig:ui}.A) and by the initial knowledge base for PennAI. 
In production mode, the knowledge base is seeded with approximately 1 million ML results generated on 165 open-source datasets, detailed here~\cite{olson_data-driven_2017}.
The user may also specify their own domain-specific cache of results. 
Below, we describe several recommender strategies that are benchmarked in the experimental section of this paper. 
Most of these recommenders are adapted from the Surprise recommender library~\cite{hug_surprise_2017}.

\subsubsection{Neighborhood Approaches}
We test four different neighborhood approaches to recommending algorithm configurations that vary in their definitions of the neighborhood. 
Three of these implementations are based the $k$-nearest neighbors (KNN) algorithm, and the other uses co-clustering. 
For each of the neighborhood methods, similarity is calculated using the mean squared deviation metric. 

In the first and second approach, clusters are derived from the results data directly and used to estimate the ranking of each ML method by computing the centroid of rankings within the neighborhood. 
Let $N_d^k(a)$ be the $k$-nearest neighbors of algorithm configuration $a$ that have been run on dataset $d$.
For \textbf{KNN-ML}, we then estimate the ranking from this neighborhood as:
\begin{equation}
    \hat{r}_{ad} = \frac{\sum_{b \in N_d^k(a)}{sim(a,b) \cdot {r}_{bd}}}{\sum_{b \in N_d^k(a)}{sim(a,b)}} 
    \label{eq:knnml}
\end{equation}

For \textbf{KNN-data}, we instead define the neighborhood over datasets, with $N_a^k(d)$ consisting of the $k$ nearest neighbors to dataset $d$ that have results from algorithm $a$. 
Then we estimate the rating as: 
\begin{equation}
    \hat{r}_{ad} = \frac{\sum_{e \in N_a^k(d)}{sim(d,e) \cdot {r}_{ae}}}{\sum_{e \in N_a^k(d)}{sim(d,e)}} 
    \label{eq:knndata}
\end{equation}

Instead of choosing to define the clusters according to datasets or algorithms, we may define co-clusters to capture algorithms and datasets that cluster together. 
This is the motivation behind co-clustering~\cite{george_scalable_2005}, the third neighborhood-based approach in this study. 
Under the \textbf{CoClustering} approach, the rating of an algorithm configuration is estimated as:
\begin{equation}
    \hat{r}_{ad} = \bar{\mathcal{C}}_{ad} + (\mu_a - \bar{\mathcal{C}}_a) + (\mu_d - \bar{\mathcal{C}}_d)
        \label{eq:cocluster}
\end{equation}
where $\bar{\mathcal{C}}$ is the average rating in cluster $\mathcal{C}$.  
As Eqn.~\ref{eq:cocluster} shows, clusters are defined with respect to $a$ and $d$ together and separately. 
Co-clustering uses a $k$-means strategy to define these clusters. 
In case the dataset is unknown, the average algorithm rating, $\mu_a$, is returned instead; likewise if the algorithm configuration is unknown, the average dataset rating $\mu_d$ is used. 
In case neither is known, the global average rating $\mu$ is returned. 

Finally, we test a metalearning method dubbed \textbf{KNN-meta}.
In this case, the neighborhood is defined according to metafeature similarity, in the same way as other approaches~\cite{brazdil_ranking_2003,brazdil_metalearning:_2008, feurer_efficient_2015}. 
We use a set of 45 metafeatures calculated from the dataset, including properties such as average correlation with the dependent variable; statistics describing the mean, max, min, skew and kurtosis of the distributions of each independent variable; counts of types of variables; and so on.

Rather than attempting to estimate ratings of every algorithm, KNN-meta maintains an archive of the best algorithm configuration for each dataset experiment. 
Given a new dataset, KNN-meta calculates the $k$ nearest neighboring datasets and recommends the highest scoring algorithm configurations from each dataset. 
KNN-meta has the advantage in cold starts since it does not have to have seen a dataset before to reason about its similarity to other results; it only needs to know how its metafeatures compare to previous experiments. 
KNN-meta has the limitation, however, that it can only recommend algorithm configurations that have been tried on neighboring datasets. 
In the case that all of these algorithm configurations have already been recommended, KNN-meta will recommends uniform-randomly from algorithms and their configurations.

\subsubsection{Singular Value Decomposition}
The singular value decomposition (\textbf{SVD}) recommender is a CF method popularized by the top entries to the Netflix challenge~\cite{bennett_netflix_2007}. 
Like other top entrants~\cite{bell_bellkor_2007,bell_scalable_2007}, SVD is based on a matrix factorization technique that attempts to minimize the error of the rankings via stochastic gradient descent (SGD). 
Each rating is estimated as
\begin{equation}
    \hat{r} = \mu + b_d + b_a + \mathbf{q}_a^T\mathbf{p}_d
    \label{eq:svd_rating}
\end{equation}

where $\mu$ is the average score of all datasets across all learners; $b_a$ is the estimated bias for algorithm $a$, initially zero; $b_d$ is the estimated bias for dataset $d$, initially zero;  $\mathbf{q}_a$ is a vector of factors associated with algorithm $a$ and $\mathbf{p}_d$ is the vector of factors associated with dataset $d$, both initialized from normal distributions centered at zero. 
Ratings are learned to minimize the regularized loss function
\begin{equation}
    \mathcal{L} =     \sum_{r_{ad} \in \mathcal{D}} \left(r_{ad} - \hat{r}_{ad} \right)^2 +         
        \lambda\left(b_a^2 + b_d^2 + ||\mathbf{q}_a||^2 + ||\mathbf{p}_d||^2\right)
\end{equation}

One of the attractive aspects of SVD is its ability to learn latent factors of datasets and algorithms ($q_d$ and $q_d$) that interact to describe the observed experimental results without explicitly defining these factors, as is done in metalearning. 
A historical challenge of SVD is its application to large, sparse matrices, such as the matrix defined by datasets and algorithms (in our experiments this matrix is about 1 million elements). 
SVD recommenders address the computational hurdle by using SGD to estimate the parameters of Eqn.~\ref{eq:svd_rating} using available experiments (i.e. dataset ratings of algorithms) only~\cite{gorrell_generalized_2006}.
 SGD is applied by the following update rules:

\begin{eqnarray}
    \centering
    b_d &\leftarrow b_d &+ \gamma (e_{ad} - \lambda b_d)\label{eq:sgd}\\ 
    b_a &\leftarrow b_a &+ \gamma (e_{ad} - \lambda b_a) \nonumber\\
    \mathbf{p}_d &\leftarrow \mathbf{p}_d &+ \gamma (e_{ad} \mathbf{q}_a - \lambda \mathbf{p}_d)\nonumber\\
    \mathbf{q}_a &\leftarrow \mathbf{q}_a &+ \gamma (e_{ad} \mathbf{p}_d - \lambda \mathbf{q}_a)\nonumber
\end{eqnarray}
where $e_{ad} = r_{ad} - \hat{r}_{ad}$.
To facilitate online learning, the parameters in Eqn.~\ref{eq:sgd} are maintained between updates to the experimental results, and the number of iterations (epochs) of training is set proportional to the number of new results. 

\subsubsection{Slope One}

Slope one~\cite{lemire_slope_2007} is a simple recommendation strategy that models algorithm performance on a dataset as the average deviation of the performance of algorithms on other datasets with which the current dataset shares at least one analysis in common. 
To rate an algorithm configurations $a$ on dataset $d$, we first collect a set $\mathcal{R}_a(d)$ of algorithms that have been trained on $d$ \textit{and} share at least one common dataset experiment with $a$.
Then the rating is estimated as
\begin{equation}
    \hat{r}_{ad} = \mu_d + \frac{1}{\mathcal{R}_a(d)} \sum_{b \in \mathcal{R}_a(d)}{dev(a,b)}
    \label{eq:slopeone}
\end{equation}

\subsubsection{Benchmark Recommenders}
As a control, we test two baseline algorithms: a random recommender and an average best recommender. 
The \textbf{Random} recommender chooses uniform-randomly among ML methods, and then uniform-randomly among hyperparameters for that method to make recommendations.
The \textbf{Average} recommender keeps a running average of the best algorithm configuration as measured by the average validation balanced accuracy across experiments. 
Given a dataset request, the Average recommender recommends algorithm configurations in order of their running averages, from best to worst.

\end{methods}
\section{Experiments}
\label{s:experiments}

\begin{figure}
    \centering
    \includegraphics[width=\mycolwidth]{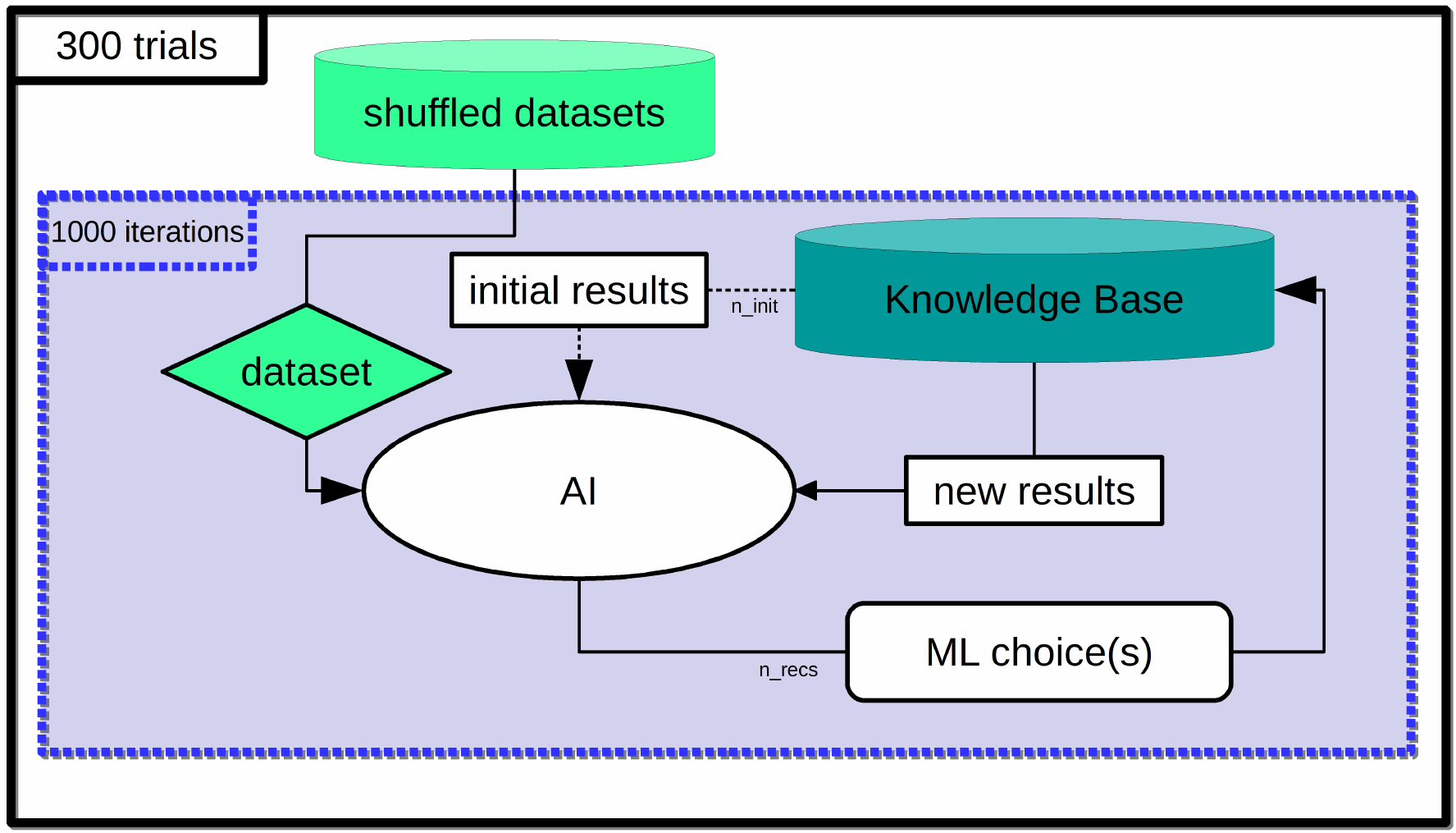}
    \caption{Diagram describing the experimental design.
    We conduct 300 trials of the experiment, each of which uses a different sampling of datasets. 
    For each trial, 1000 iterations are conducted. 
    The AI (recommender system) is initially trained on \textit{n\_init} ML results from the Knowledge Base. 
    Then, each iteration, \textit{n\_recs} recommendations are made by the AI for one dataset. 
    These recommended ML configurations are retrieved from the Knowledge Base and used to update the AI for the next iteration.
    }
    \label{fig:exp_diagram}
\end{figure}

The goal of our experiments is to assess different recommendation strategies in their ability to recommend better algorithm configurations for various datasets as they learn over time from previous experiments.
The diagram in Fig.~\ref{fig:exp_diagram} describes the experimental design used to evaluate recommendation strategies under PennAI. 

\paragraph{Datasets}
We assessed each recommender on 165 open-source classification datasets, varying in size (hundreds to millions of samples) and origin (bioinformatics, economics, engineering, etc).
We used datasets from the Penn Machine Learning Benchmark (PMLB)~\cite{olson_pmlb:_2017}. 
PMLB is a curated and standardized set of hundreds of open source supervised machine learning problems from around the web (sharing many in common with OpenML~\cite{vanschoren_openml:_2014}). 
In previous work, we conducted an extensive benchmarking experiment to characterize the performance of different ML frameworks on these problems~\cite{olson_pmlb:_2017,olson_data-driven_2017}. 
The benchmark assessed 13 ML algorithms over a range of hyperparameters detailed in Table 1 of the Supplemental Material on these problems. 
This resulted in a cache of over 1 million ML results across a broad range of classification problems which we use here to assess the performance of each recommender with a known ranking of algorithms for each dataset. 
For the experiment in this paper, we used a subset of these results consisting of 12 ML algorithms (dropping one of three na\"{i}ve Bayes algorithms) with 7580 possible hyperparameter configurations.

\paragraph{Evaluation of Recommender Systems}
\label{s:exp_rs}
The first experiment consisted of 300 repeat trials for each recommender. 
In each trial, the recommender began with a knowledge base of $n_{init}$ experiments that consist of single ML runs on single datasets. 
For each iteration of the experiment, the recommender was asked to return $n_{recs}$ recommendations for a randomly chosen dataset. 
Once the recommendation was made, the 5-fold CV results on training data for the recommended algorithm configurations were fed back to the recommender as updated knowledge. 
Note that this experiment mirrors a reinforcement learning experiment in which the actions taken by the recommender (i.e., the recommendations it makes) determine the information it is able to learn about the relationship between datasets and algorithm configurations. 
To assess the quality of these recommendations without overfitting, we used a separate, hold-out test set performance score for each algorithm configuration on each dataset.
We report the scores of each algorithm configuration on these heldout data throughout Section~\ref{s:results}.
For our experiments, we varied $n_{init} \in [1, 100, 10000]$ and $n_{recs} \in [1, 10, 100]$.
These settings control sensitivity to 1) starting with more information and 2) exploring more algorithm options during learning. 

\subsection{Comparison to State-of-the-Art}

Based on the results of our first experiment (Section~\ref{s:exp_rs}), we chose a final configuration for PennAI and benchmarked its performance against two other state-of-the-art AutoML tools: AutoSklearn~\cite{feurer_efficient_2015} and HyperOptSklearn~\cite{komer_hyperopt-sklearn:_2014}. 
For AutoSklearn, we restricted the search space to ML configurations to bring it closer to the search spaces of PennAI and HyperOptSklearn that do not use feature preprocessors.
Otherwise we used default settings of both AutoSklearn and HyperOptSklearn. 
For this comparison, we performed a leave-one-out style analysis, meaning that PennAI is trained on results from all other datasets prior to iteratively making recommendations for a given dataset. 
This leave-one-out analysis corresponds to the applied case, in which PennAI is deployed with a pre-trained recommender and must run experiments for a newly uploaded dataset. 
For each method, we assessed the generalization performance of the returned model after a given number of evaluations.
For AutoSklearn, we assessed its performance as a function of wall-clock run-time, since there was not an apples-to-apples way to control the number of algorithm evaluations.

\paragraph{Comparison Metrics}
Since we have the complete results of ML analyses on our experiment datasets, we assessed recommendations in terms of their closeness to the best configuration, i.e. that configuration with the best holdout performance on a dataset among all algorithms in our exhaustive benchmark. 
Each algorithm configuration is primarily assessed according to its \textit{balanced accuracy} ($BA$), a metric that takes into account class imbalance by averaging accuracy over classes\cite{velez_balanced_2007}.  
Let the best balanced accuracy on a given dataset be $BA_d^*$. 
Then the performance of a recommendation is assessed as the relative distance to the best solution, as:
\begin{equation}
    \Delta \text{Balanced~Accuracy}_{ad} = \frac{(BA_d^*-BA_{ad})}{BA_d^*} \label{eq:dba}
\end{equation}

In addition to Eqn.~\ref{eq:dba}, we assessed the AI in terms of the number of datasets for which it is able to identify an ``optimal" configuration. 
Here we defined ``optimal" algorithm configurations to be those that score within some small threshold of the best performance for a dataset.   
This definition of optimality is of course limited, both by the finite search space defined by the algorithm configuration options and by the choice of thresholding (we tried 1\% and 5\%).
Nonetheless, this definition gives a practical indicator of the extent to which AI is able to reach the best known performance. 

\subsection{Illustrative Example}
\label{s:ill_ex}
In addition to testing different recommendation strategies within PennAI, we applied PennAI to the task of generating a classification model for predicting patient's risk of septic shock. 
For the shock task, we used data from the MIMIC-III~\cite{johnson_mimic-iii_2016} critical care database and preprocessed it according to prior work~\cite{harutyunyan_multitask_2019}. 
In addition to the binning process described by Harutyunyan et al., we calculated autocorrelations for each predictor at 5 different lags in order to capture time series features. 
This resulted in a prediction problem with 8346 training patients, 6284 test patients, and 60 dependent variables. 

We used the best performing PennAI configuration from our experiments.
We began by allowing the AI to suggest and run 10 experiments. 
We then manually chose 5 additional algorithm configurations to run, using default settings in PennAI.
PennAI's results page was studied to validate models on the training set, and to pick a final model for download. 
We then evaluated this downloaded model on the testing set using PennAI's model export functionality.
Screenshots of this process are shown in Fig.~\ref{fig:ui}.

As a point of comparison, we trained a long-term-short-term memory (LSTM) deep learning model using the architecture from~\cite{harutyunyan_multitask_2019} and training for 100 epochs. 
We also compared the results to septic shock models from literature~\cite{henry_targeted_2015}.
We discuss these results in Section~\ref{s:results_ill}.

\section{Results}
\label{s:results}

\begin{figure*}[ht!]
    \centering
    \includegraphics[width=0.7\textwidth]{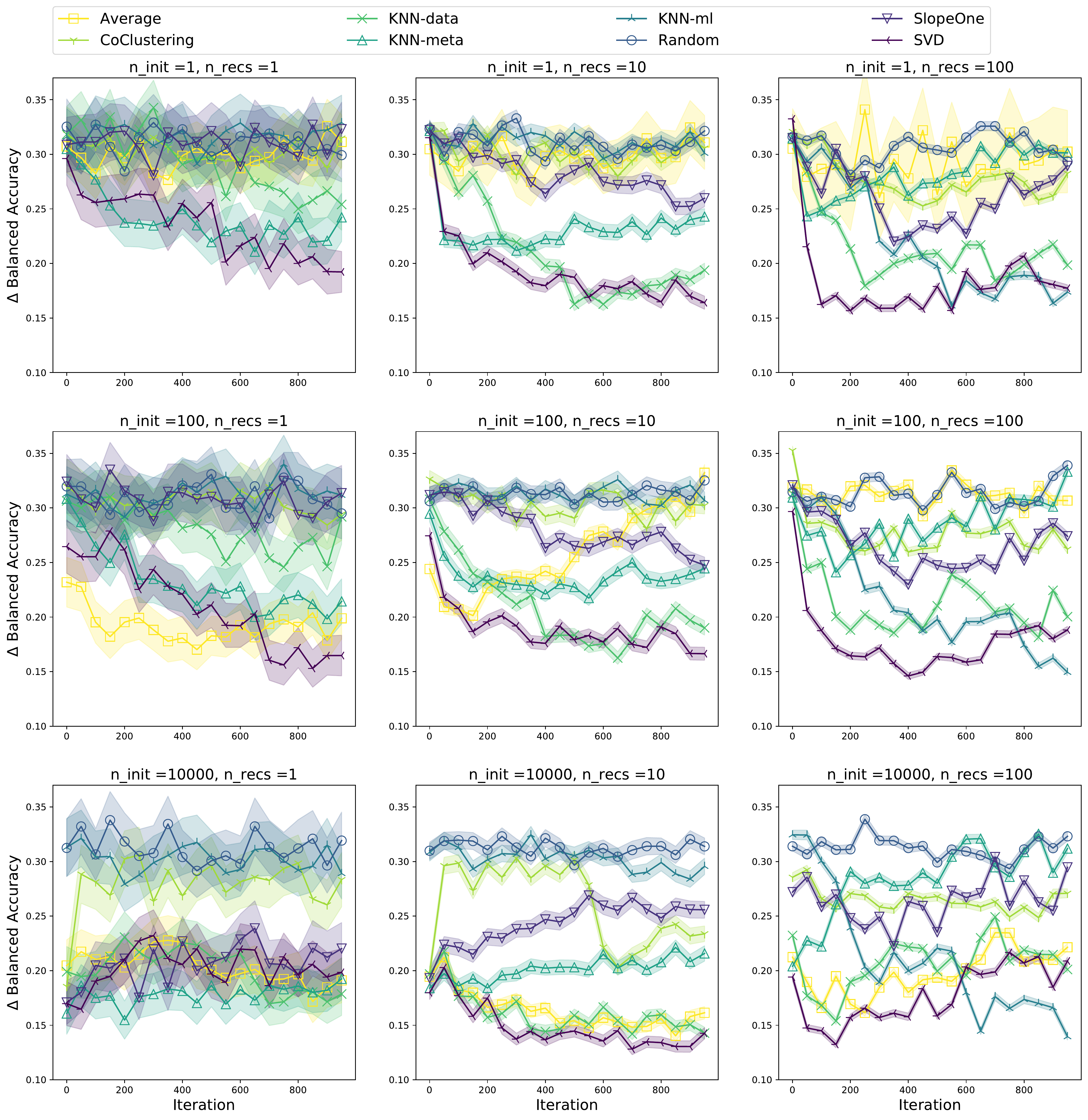}
    \caption{
        Experiment results for each recommendation strategy. 
        Each plot shows the median $\Delta$ Balanced Accuracy (Eqn.~\ref{eq:dba} for 300 trials with error bars denoting 95\% confidence intervals. 
    From left to right, the number of recommendations made per dataset increases; from top to bottom, the number of experiments in the initial knowledgebase increases.
    A lower $\Delta$ Balanced Accuracy indicates that ML configurations being recommended are closer to the best known configuration.
    }
    \label{fig:exp}
\end{figure*}

Results for the PMLB experiment are shown in Figure~\ref{fig:exp}. 
Recommenders are first compared in terms of median $\Delta$ Balanced Accuracy (Eq.~\ref{eq:dba}) in Fig.~\ref{fig:exp}. 
In Fig.~\ref{fig:cum_success}, we look at the fraction of datasets for which SVD is able to find an optimal configuration under different experiment treatments. 
In the Supplemental Material, we visualize the behavior of a subset of recommenders in order to gain insight into which algorithms are being selected and how this compares to the underlying distribution of algorithm performance in the knowledgebase.

Let us first focus on the performance results in Fig.~\ref{fig:exp}. 
We find in general that the various recommender systems are able to learn to recommend algorithm configurations that increasingly minimize the gap between the best performance on each dataset (unkown to the AI) and the current experiments.  
SVD performs the best, tending to reach good performance more quickly than the other recommendation strategies, across treatments. 
KNN-data and KNN-ml are the next best methods across treatments; KNN-ML shows a sensitivity to the number of recommendations per iteration, indicating it requires more results to form good clusters. 
For most experimental treatments, is a gap between those three methods and the next best recommenders, which vary between SlopeOne, KNN-meta, and CoClustering for different settings.

As we expected due to its cold-start strategy, KNN-meta turns out to work well initially, but over time fails to converge on a set of high quality recommendations.
The collaborative filtering recommendaters are generally able to learn quickly from few examples compared to the metalearning approach. 
This difference in performance has been found in other domains, particularly movie recommendations~\cite{pilaszy_recommending_2009}.

Given 100 initial experiment results and 10 recommendations per iteration, the SVD recommender converges to within 5\% of the optimal performance in approximately 100 iterations, corresponding to approximately 7 training instances per dataset. 
Note that the performance curves begin to increase on the right-most plots that correspond to 100 recommendations per dataset. 
In these cases, the recommender begins recommending algorithms configurations with lower rankings due to repeat filtering, described in Section~\ref{s:methods}.

\subsection{Comparison to AutoML}


\begin{figure*}
    \begin{minipage}{\mycolwidth}
    \centering
    \includegraphics[width=\textwidth]{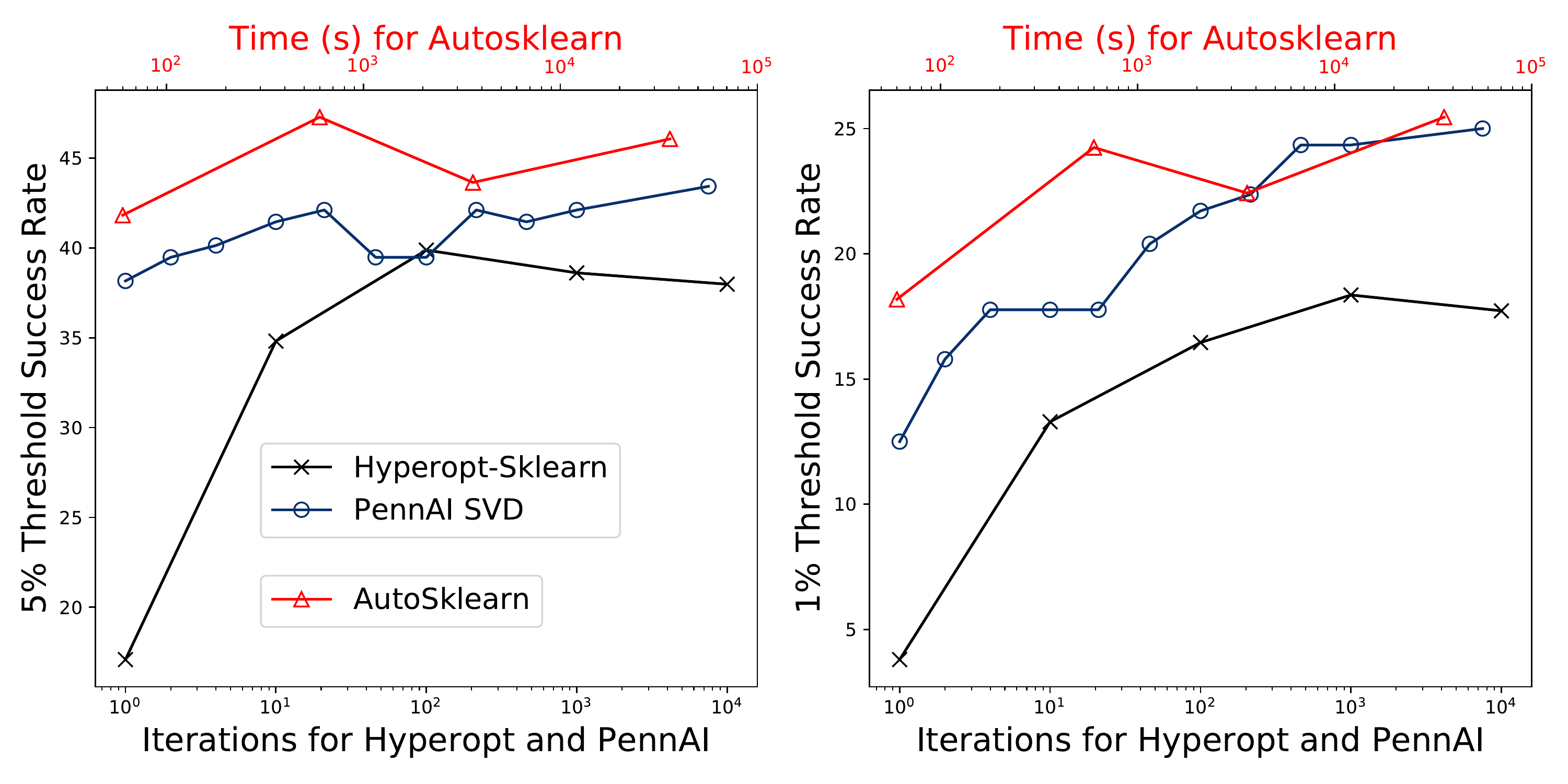}
    \caption{
        Cumulative success rates across all datasets using PennAI, AutoSklearn, and HyperOptSklearn. 
        The success rate is the fraction of datasets for which the recommender has trained an algorithm configuration that achieves a test set balanced accuracy within 1 or 5\% of the best performance on that dataset. 
    }
    \label{fig:cum_success}
\end{minipage}
\hfill
\begin{minipage}{\mycolwidth}

    \centering
    \includegraphics[width=0.8\textwidth]{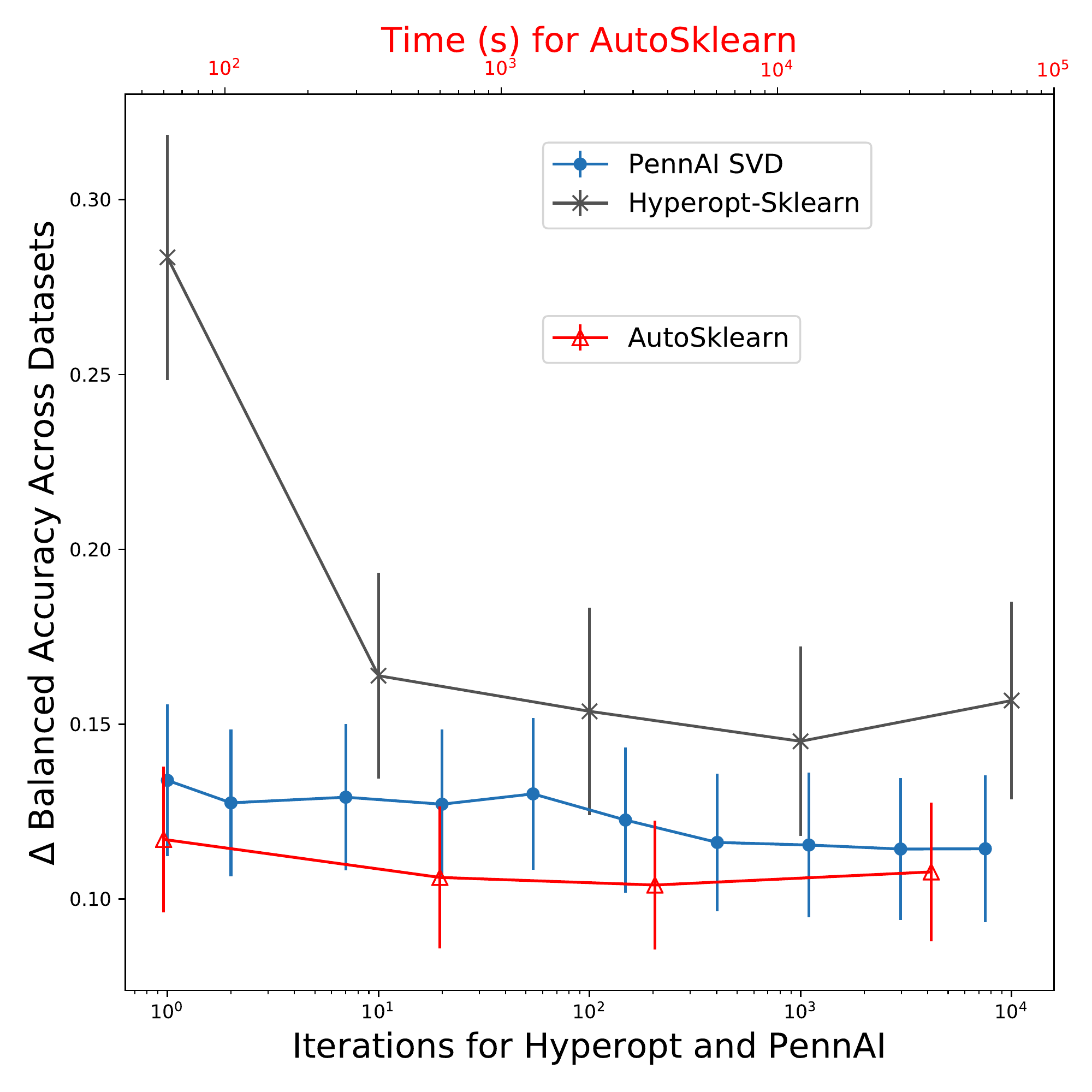}
    \caption{
        $\Delta$BA across all datasets, using PennAI, AutoSklearn, and HyperOptSklearn. 
        Error bars indicate the 95\% confidence intervals.
    }
    \label{fig:dba}
\end{minipage}
\end{figure*}


In Fig.~\ref{fig:cum_success} and ~\ref{fig:dba}, we compare PennAI with SVD to AutoSklearn and Hyperopt, two widely used AutoML tools. 
We choose to use the SVD algorithm for PennAI since it is shown to perform the best in terms of $\Delta$ Balanced Accuracy in our prior analysis. 
The side-by-side graphs of Fig.~\ref{fig:cum_success} show the percent of datasets for which a given method is able to return an algorithm within 5\% (left) and 1\% (right) of the best performance, as a function of computational effort. 
The results show AutoSklearn finding best configurations most often for the 5\% threshold, and AutoSklearn and PennAI finding best configurations at about the same rate for 1\% threshold. 
In both cases, the performance of AutoSklearn and PennAI differs only by a few percent, whereas HyperOptSklearn tends to under-perform those two methods. 
Fig.~\ref{fig:dba} considers the same results without a threshold, instead showing the average reduction in $\Delta$ Balanced Accuracy (Eqn.~\ref{eq:dba}) across datasets for each method. 
By this metric, AutoSklearn performs the best, although its performance is not significantly different than that of PennAI, as observed by the overlap of their error bars.

\subsection{Illustrative Example}
\label{s:results_ill}

The screenshots in Fig.~\ref{fig:ui} show the steps in the septic shock model fitting procedure. 
In Table~\ref{tbl:models}, the fitted models are detailed, including the ML configuration, 5-fold CV AUROC score. 
Based on these scores, the gradient boosting model shown in bold was selected and exported from PennAI. 
This model achieved an AUROC of 0.85 $\pm$ 0.02 on the test data, compared to a mean of 0.86 $\pm$ 0.02 for the LSTM model trained for 100 epochs. 
An advantage of the PennAI-generated model is that it took less than 2 minutes to train, whereas the LSTM model took more than 30 hours, using an NVIDIA GeForce GTX 970. 
Two caveats of this time comparison are that 1) the LSTM is multi-task, i.e. it makes predictions for 25 different phenotypes; and 2) the results were trained on different hardware. 
The PennAI results were generated on a single thread, Intel(R) Core(TM) i7-6950X CPU @ 3.00GHz. 
The LSTM results were generated using an NVIDIA GeForce GTX 970 graphics processing unit. 
Nevertheless, the training time difference of 915x is substantial. 

\begin{figure}
    \begin{minipage}{\mycolwidth}
    \centering
        \includegraphics[width=\textwidth]{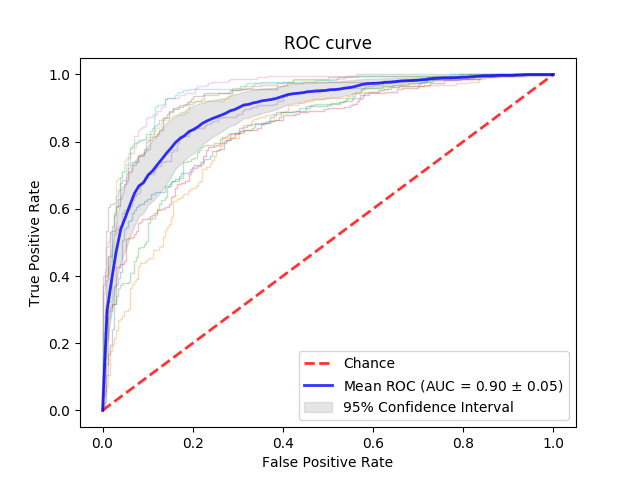}
    \end{minipage}
    \\
    \begin{minipage}{\mycolwidth}
        \includegraphics[width=\textwidth]{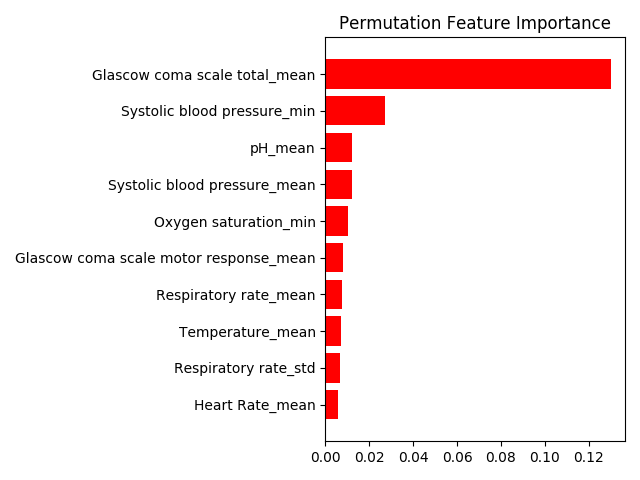}
    \end{minipage}
    \caption{
        Top: ROC curve for five validation folds of the final septic shock model.
        Bottom: Permutation importance values for dataset features, with larger values indicating features that are more important for prediction.
    }
    \label{fig:shock}
\end{figure}

Both model performances are in a similar range to state-of-the-art early detection systems recently deployed to identify septic shock in critical care patients~\cite{henry_targeted_2015}. 
Figure~\ref{fig:shock} shows the cross-validation ROC curves for the gradient boosting model of shock, as well as a sensitivity analysis of the final model. 
The two most important factors to prediction are the mean Glasgow coma scale rating for the patient and their minimum systolic blood pressure reading. 
The importance of these two factors lends credence to the model, since they are important for assessing septic shock~\cite{henry_targeted_2015}.  
The Glasgow coma scale is an indicator of assessment of the patient's consciousness and therefore a likely indicator for adverse events. 
A drop in systolic blood pressure is a tell-tale signature of septic shock and is used as a clinical diagnostic~\cite{singer_third_2016}.

\begin{table*}
    \scriptsize
    \caption{Fitted models chosen by the user and AI for predicting shock. 
             Area under the receiver operating curve (AUROC) scores are reported with 95\% confidence intervals. 
         For the selected model shown in bold, we report AUROC on the test set, which is within 1\% of state-of-the-art results for this task (LSTM, bottom row).}\label{tbl:models}
    \begin{tabularx}{\textwidth}{l l l r r r} \toprule
        Model   &   Hyperparameters (Sklearn syntax)            &  Source   &   5-fold CV AUROC &  Test AUROC   &   Training time\\ \midrule
        Logistic Regression &   penalty=L2, C=1.0, Dual=False   
                            &   user     &   0.84 $\pm$ 0.03     &   -  &   13 s    \\ 
                            &   penalty=L1, C=1e-3, Dual=False   
                            &   AI  &      0.81 $\pm$ 0.03       &  -   &   16 s    \\ 
        KNN Classifier      &   n\_neighbors=1, weights=`distance'  
                            &   AI  &   0.68 $\pm$ 0.02     &   -   &   1 m 1 s     \\
                            &   n\_neighbors=7, weights=`uniform'  
                            &   AI  &   0.81 $\pm$ 0.03     &   -   &   11 m 3 s    \\
                            &   n\_neighbors=11, weights=`uniform'  
                            &   AI  &   0.82 $\pm$ 0.03     &   -   &   11 m 13 s   \\
        Support Vector      &   kernel=rbf, C=1.0, degree=3, gamma=0.01 
                            &   user    &   0.70    $\pm$ 0.03      &   -   &   7 m 55 s    \\
                            &   kernel=polynomial, C=1.0, degree=3, gamma=0.01 
                            &   user    &   0.68    $\pm$ 0.03  &   -   &   26 m 8 s \\
        Classification Tree &   criterion=entropy, max\_depth=10
                            &   AI      &   0.75 $\pm$ 0.02 &   -   &   5 s         \\
                            &   criterion=gini, max\_depth=10
                            &   user      &   0.73 $\pm$ 0.06 &   -   & 6 s         \\
        Random Forest       &   criterion=Gini, max\_features=`sqrt', n\_estimators=100
                            &   user    &  0.88 $\pm$ 0.06  &   -   &   2 m 7 s     \\ 
                            &   criterion=entropy, max\_features=`log2', n\_estimators=100
                            &   AI &  0.83 $\pm$ 0.06  &   -   &    1 m 16 s         \\ 
        {\bf Gradient Boosting}   &   {\bf learning\_rate=0.01, max\_depth=5, n\_estimators=500}
                                  &   {\bf AI}       &   {\bf 0.90 += 0.05}    &   {\bf 0.85 $\pm$0.02    }     &   \textbf{1 m 58 s}    \\
                            &   learning\_rate=0.1, max\_depth=5, n\_estimators=500    
                            &   AI       &   0.89 += 0.05    &   -  &   2 m 25 s    \\
                            &   learning\_rate=0.1, max\_depth=3, n\_estimators=500    
                            &   AI       &   0.89 += 0.05    &   -   &  1 m 18 s    \\
                            &   learning\_rate=1, max\_depth=1, n\_estimators=100    
                            &   AI       &   0.88 += 0.06    &   -   &  10 s    \\
        \textit{LSTM}                &   \textit{multi-task, 100 epochs}    &   \textit{Harutyunyan et al.\cite{harutyunyan_multitask_2019}} &  -   &   \textit{0.86 $\pm$ 0.02 }    &  \textit{30 h 21 m} \\
        \bottomrule
    \end{tabularx}
\end{table*}

\section{Conclusion}
\label{s:conclusion}

In this paper we propose a data science tool for non-experts that generates increasingly reliable ML recommendations tailored to the needs of the user. 
The web-based interface provides the user with an intuitive way to launch and monitor analyses, view and understand results, and download reproducible experiments and fitted models for offline use. 
The learning methodology is based on a recommendation system that can learn from both cached and generated results.
We demonstrate through the experiments in this paper that collaborative filtering algorithms can successfully learn to produce intelligent analyses for the user starting with sparse data on algorithm performance. 
We find in particular that a matrix factorization algorithm, SVD, works well in this application area. 

PennAI automates the algorithm selection and tuning problem using a recommendation system that is bootstrapped with a knowledgebase of previous results. 
The default knowledgebase is derived from a large set of experiments conducted on 165 open source datasets. 
The user can also configure their own knowledgebase of datasets and results catered to their application area. 
In our application example, we used PennAI with a generic knowledgebase of datasets to successfully train and validate a predictive model of septic shock; we found that PennAI was able to quickly suggest a state-of-the-art model, with little user input. 
In the future, we hope to further improve PennAI's ability to handle domain-specific tasks by creating knowledgebases for particular areas such as electronic health records and genetics. 

We also hope that PennAI can serve as a testbed for novel AutoML methodologies. 
In the near term we plan to extend the methodology in the following ways. 
First, we plan to implement a focused hyperparameter tuning strategy that can fine-tune the models that are recommended by the AI, similar to AutoSklearn \cite{feurer_efficient_2015} or Hyperopt~\cite{komer_hyperopt-sklearn:_2014}.
We plan to make this process transparent to the user so that they may easily choose which models to tune and for how long.
We also plan to increasingly automate the data preprocessing, which is, at the moment, mostly up to the user. 
This can include processes from imputation and data standardization to more complex operations like feature selection and engineering. 

\section{Acknowledgments}

The authors would like thank the members of the Institute for Biomedical Informatics at Penn for their many contributions. 
Contributors included Sharon Tartarone, Josh Cohen, Randal Olson, Patryk Orzechowski, and Efe Ayhan. 
We also thank John Holmes, Moshe Sipper, and Ryan Urbanowicz for their useful discussions.
The development of this tool was supported by National Institutes of Health research grants (K99 LM013256-01, R01 LM010098 and R01 AI116794) and National Institutes of Health infrastructure and support grants (UC4 DK112217, P30 ES013508, and UL1 TR001878).




\section*{Supplementary Material}

\section*{System Architecture}

\begin{figure}
    \centering
    \includegraphics[width=\textwidth]{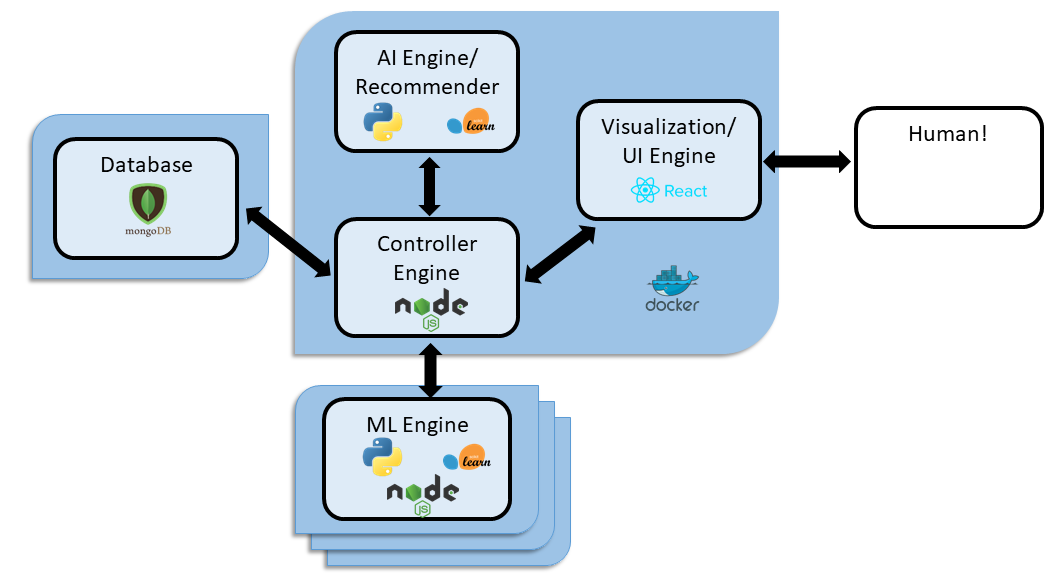}
    \caption{Diagram describing the architecture of PennAI.}
    \label{fig:arch}
\end{figure}

PennAI is a multi-component architecture that uses a variety of technologies including Docker\footnote{\url{https://www.docker.com/}}, Python, Node.js\footnote{\url{https://nodejs.org}}, scikit-learn\footnote{\url{http://sklearn.org}}, FGLab~\footnote{\url{https://kaixhin.github.io/FGLab/}}, and MongoDb\footnote{\url{https://www.mongodb.com/}}.  
The architecture is shown in Figure~\ref{fig:arch}.
The project contains multiple docker containers that are orchestrated by a docker-compose file.  
The central component is the controller engine, a server written in Node.js.  
This component is responsible for managing communication between the other components using a REST API.  
A MongoDb database is used for persistent storage.  
The UI component is a web application written in javascript that uses the React library to create the user interface and the Redux library to manage UI state.  
The interface supports user interactions including uploading datasets for analysis, requesting AI recommendations for a dataset, manually specifying and running machine learning experiments, and displaying experiment results in an intuitive way.  
The AI engine is written in Python.  
As users make requests to perform analysis on datasets, the AI engine will generate new machine learning experiment recommendations and communicate them to the controller engine.  
The AI engine contains a knowledgebase of previously run experiments, results, and dataset metafeatures that it uses to inform the recommendations it makes.  
The knowledgebase is bootstrapped with a collection of experiment results generated from the PMBL benchmark datasets.  
Instructions and code templates are provided to allow easy integration of custom recommendation systems.  
The machine learning component is responsible for running machine learning experiments on datasets. 
It has a Node.js server that is used to communicate with the controller engine, and uses python to execute Scikit-learn experiments on datasets and communicate results back to the central server.  
A PennAI instance can support multiple instances of machine learning engines, enabling multiple experiments to be run in parallel.

\section*{Additional Experiment Details}

In Table~\ref{tbl:algos}, the parameter spaces of each ML algorithm that was used in our experimental analysis is shown. 
In total, there were 7580 combinations. 

\begin{table*}[h!]
\caption{Analyzed algorithms with their parameters settings. 
The methods and parameters are named according to Scikit-learn nomenclature\cite{pedregosa_scikit-learn:_2011}. }
\label{tbl:algos}
\footnotesize
\begin{center}
\begin{tabular}{l c c}
\textbf{Algorithm name} &\textbf{Parameter} &\textbf{Values}  \\
\toprule
\multirow{2}{*}{AdaBoostClassifier}	&	 learning\_rate 	&	 [0.01, 0.1, 0.5, 1.0, 10.0, 50.0, 100.0] \\
	&	 n\_estimators 	&	 [10, 50, 100, 500, 1000] \\
\midrule
\multirow{3}{*}{BernoulliNB}	&	 alpha 	&	 [0.0, 0.1, 0.25, 0.5, 0.75, 1.0, 5.0, 10.0, 25.0, 50.0] \\
	&	 fit\_prior 	&	 ['true', 'false'] \\
	&	 binarize 	&	 [0.0, 0.1, 0.25, 0.5, 0.75, 0.9, 1.0] \\
\midrule
\multirow{3}{*}{DecisionTreeClassifier}	&	 min\_weight\_fraction\_leaf 	&	 [0.0, 0.05, 0.1, 0.15, 0.2, 0.25, 0.3, 0.35, 0.4, 0.45, 0.5] \\
	&	 max\_features 	&	 [0.1, 0.25, 0.5, 0.75, 'log2', None, 'sqrt'] \\
	&	 criterion 	&	 ['entropy', 'gini'] \\
\midrule
\multirow{4}{*}{ExtraTreesClassifier}	&	 n\_estimators 	&	 [10, 50, 100, 500, 1000] \\
	&	 min\_weight\_fraction\_leaf 	&	 [0.0, 0.05, 0.1, 0.15, 0.2, 0.25, 0.3, 0.35, 0.4, 0.45, 0.5] \\
	&	 max\_features 	&	 [0.1, 0.25, 0.5, 0.75, 'log2', None, 'sqrt'] \\
	&	 criterion 	&	 ['entropy', 'gini'] \\
\midrule
\multirow{5}{*}{GradientBoostingClassifier}	&	 loss 	&	 ['deviance'] \\
	&	 learning\_rate 	&	 [0.01, 0.1, 0.5, 1.0, 10.0] \\
	&	 n\_estimators 	&	 [10, 50, 100, 500, 1000] \\
	&	 max\_depth 	&	 [1, 2, 3, 4, 5, 10, 20, 50, None] \\
	&	 max\_features 	&	 ['log2', 'sqrt', None] \\
\midrule
\multirow{2}{*}{KNeighborsClassifier}	&	 n\_neighbors 	&	 [1, 2, $\dots$, 25] \\
	&	 weights 	&	 ['uniform', 'distance'] \\
\midrule
\multirow{4}{*}{LogisticRegression}	&	 C 	&	 [0.5, 1.0, $\dots$, 20.0] \\
	&	 penalty 	&	 ['l2', 'l1'] \\
	&	 fit\_intercept 	&	 ['true', 'false'] \\
	&	 dual 	&	 ['true', 'false'] \\
\midrule
\multirow{2}{*}{MultinomialNB}	&	 alpha 	&	 [0.0, 0.1, 0.25, 0.5, 0.75, 1.0, 5.0, 10.0, 25.0, 50.0] \\
	&	 fit\_prior 	&	 ['true', 'false'] \\
\midrule
\multirow{3}{*}{PassiveAggressiveClassifier}	&	 C 	&	 [0.0, 0.001, 0.01, 0.1, 0.5, 1.0, 10.0, 50.0, 100.0] \\
	&	 loss 	&	 ['hinge', 'squared\_hinge'] \\
	&	 fit\_intercept 	&	 ['true', 'false'] \\
\midrule
\multirow{4}{*}{RandomForestClassifier}	&	 n\_estimators 	&	 [10, 50, 100, 500, 1000] \\
	&	 min\_weight\_fraction\_leaf 	&	 [0.0, 0.05, 0.1, 0.15, 0.2, 0.25, 0.3, 0.35, 0.4, 0.45, 0.5] \\
	&	 max\_features 	&	 [0.1, 0.25, 0.5, 0.75, 'log2', None, 'sqrt'] \\
	&	 criterion 	&	 ['entropy', 'gini'] \\
\midrule
\multirow{8}{*}{SGDClassifier}	&	 loss 	&	 ['hinge', 'perceptron', 'log', 'squared\_hinge', 'modified\_huber'] \\
	&	 penalty 	&	 ['elasticnet'] \\
	&	 alpha 	&	 [0.0, 0.001, 0.01] \\
	&	 learning\_rate 	&	 ['constant', 'invscaling'] \\
	&	 fit\_intercept 	&	 ['true', 'false'] \\
	&	 l1\_ratio 	&	 [0.0, 0.25, 0.5, 0.75, 1.0] \\
	&	 eta0 	&	 [0.01, 0.1, 1.0] \\
	&	 power\_t 	&	 [0.0, 0.1, 0.5, 1.0, 10.0, 50.0, 100.0] \\
\midrule
\multirow{5}{*}{SVC}	&	 C 	&	 [0.01] \\
	&	 gamma 	&	 [0.01] \\
	&	 kernel 	&	 ['poly'] \\
	&	 degree 	&	 [2, 3] \\
	&	 coef0 	&	 [0.0, 0.1, 0.5, 1.0, 10.0, 50.0, 100.0] \\

\bottomrule
\end{tabular}
\end{center}
\end{table*}

\section{Additional Results}

\begin{figure}
    \centering
    \begin{minipage}{0.75\textwidth}
    \centering
    \includegraphics[width=\textwidth]{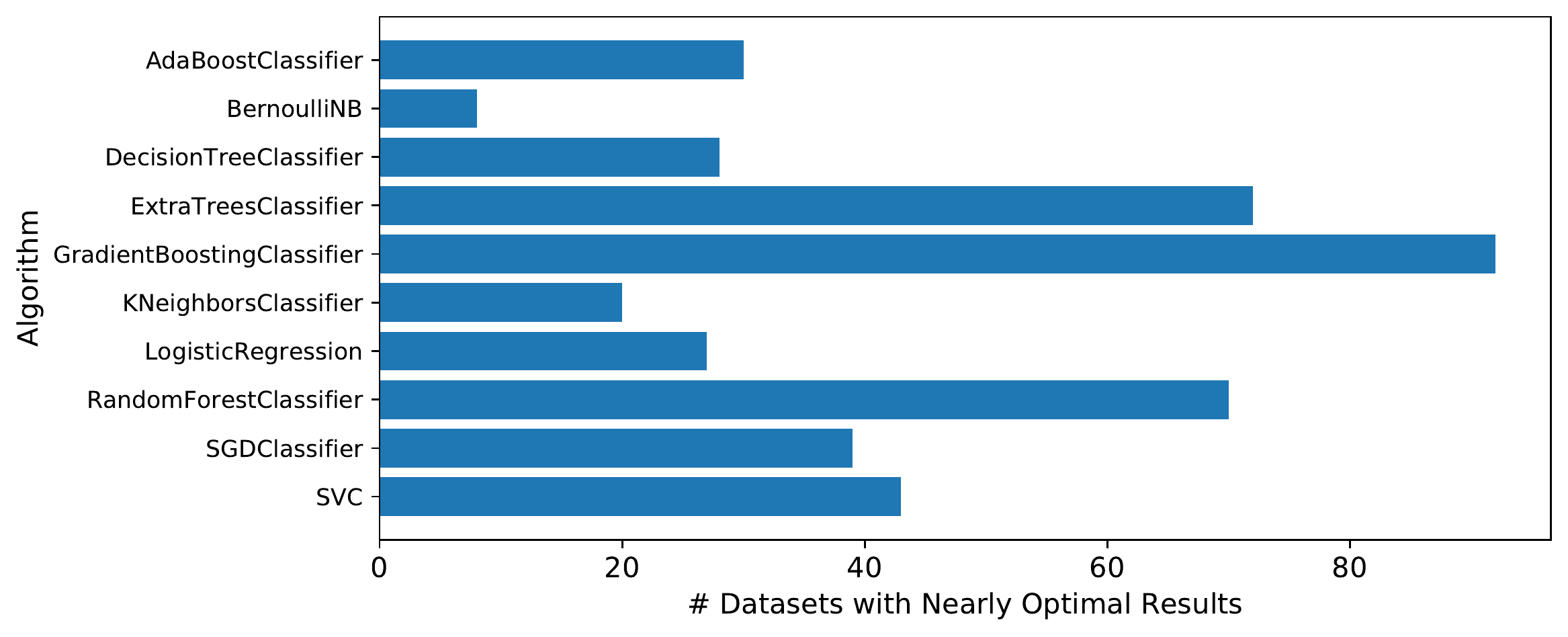}
    \end{minipage}
    \\
    \begin{minipage}{0.75\textwidth}
    \includegraphics[width=\textwidth]{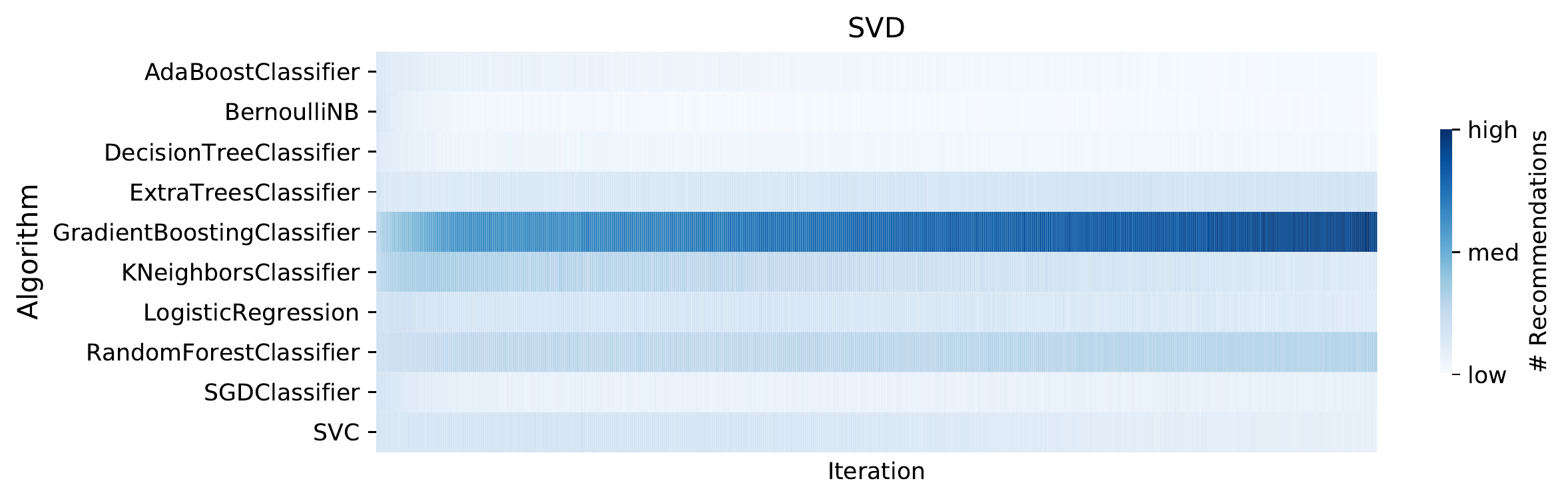}
    \end{minipage}
    \\
    \begin{minipage}{0.75\textwidth}
    \centering
    \includegraphics[width=\textwidth]{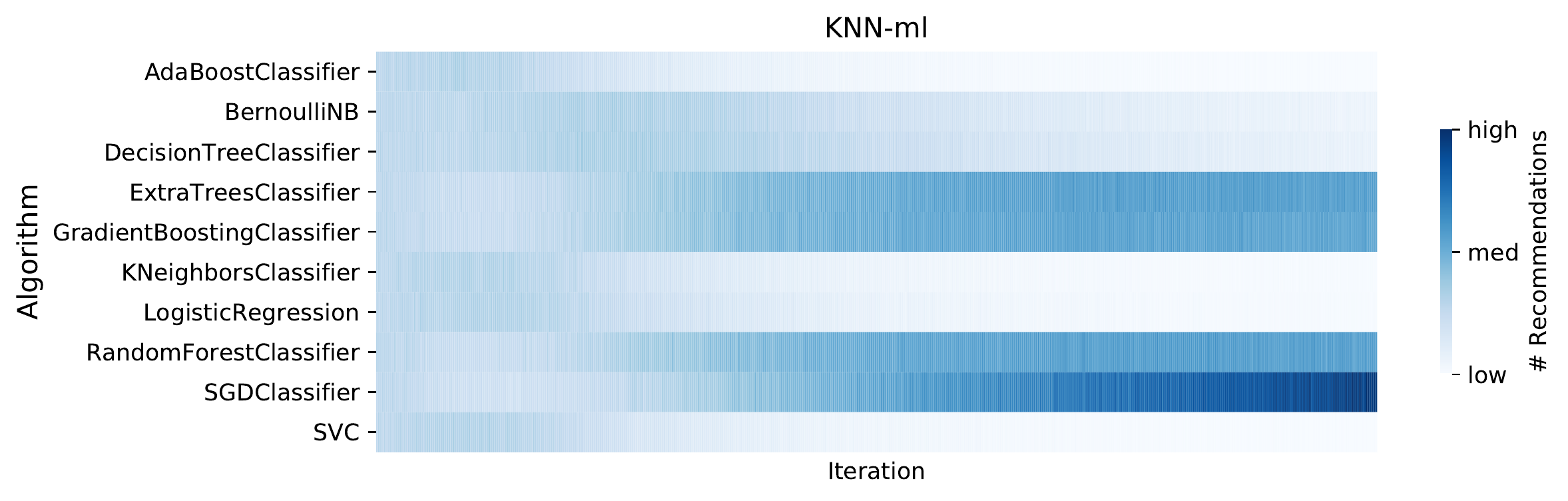}
    \end{minipage}
    \\
    \begin{minipage}{0.75\textwidth}
        \includegraphics[width=\textwidth]{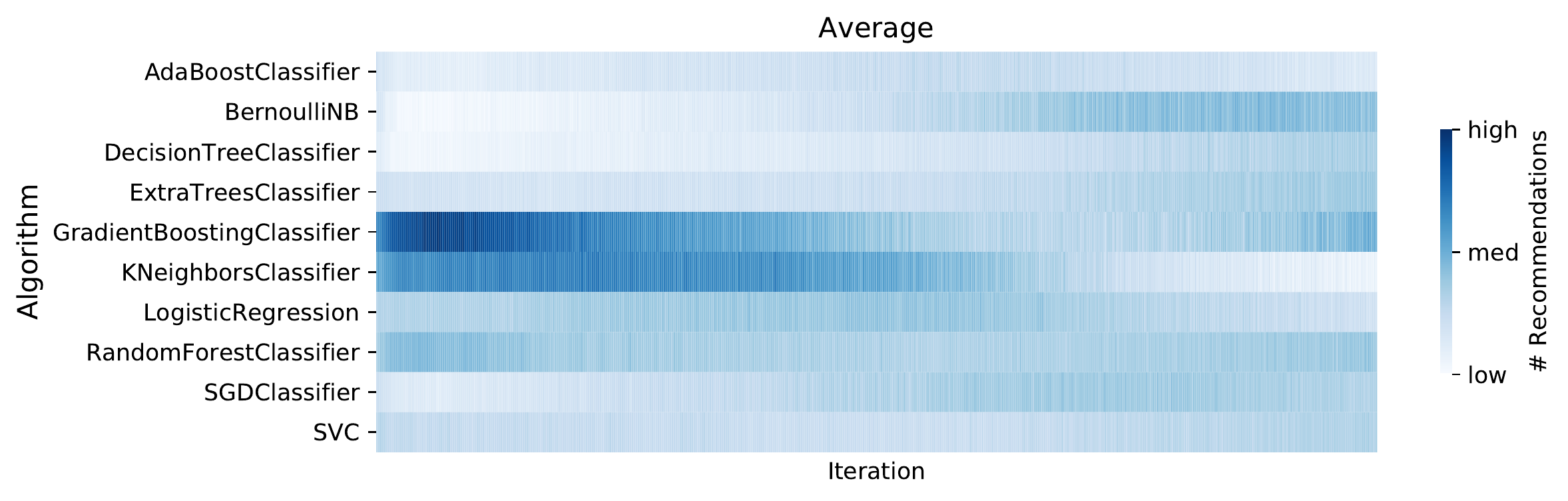}
    \end{minipage}
    \caption{
        Heatmaps of three different recommendation strategies on the PMLB benchmark showing how often each ML was recommended each iteration.
        These plots show the experiment treatment with $n_{recs} = 10$ and $n_{init} = 100$. 
        The top figure shows the number of datasets for which each ML algorithm has a configuration that is within 1\% of the best performance on that dataset. 
        The second figure shows SVD recommendations; over several iterations, it learns to approximate the frequency distribution of best ML models, i.e. GradientBoostingClassifier, followed by RandomForestClassifier and ExtraTreesClassifier. 
        The third plot shows KNN-ml recommendations; in this case, SGDClassifier ends up being recommended more often than is supported by its benchmark performance. 
        The final plot shows the performance of Average recommender; the distribution of algorithm recommendations is more wide, and tends to drift away from the best algorithm choices after several iterations. 
    }
    \label{fig:heatmap}
\end{figure}

The final set of plots in Fig.~\ref{fig:heatmap} show the frequency with which SVD, KNN-ML, and SlopeOne recommend different algorithms in comparison to the frequency of top-ranking algorithms by type (the top left plot). 
Here we see that SVD gradually learns to recommend the top five algorithms in approximately the same ranking as they appear in the knowledgebase. 
This lends some confidence to the relationship that SVD has learned between algorithm configurations and dataset performance.

\end{document}